\documentclass[conference]{IEEEtran}
\IEEEoverridecommandlockouts

\usepackage{cite}
\usepackage{amsmath,amssymb,amsfonts}
\usepackage{graphicx}
\usepackage{textcomp}
\usepackage{xcolor} 
\usepackage{times}
\usepackage{epsfig}
\usepackage{graphicx}
\usepackage{amsmath}
\usepackage{amssymb}
\usepackage[bb=dsserif]{mathalpha}
\usepackage{bm}

\usepackage{booktabs,colortbl,tabularx}
\usepackage{pifont}%

\usepackage{mathtools}
\usepackage{commath}
\usepackage{algorithm}
\usepackage{algpseudocode}

\usepackage{multirow}
\usepackage{comment} 
\usepackage{enumitem}

\setlist[itemize]{itemsep=1pt, topsep=2pt, partopsep=0pt}

\definecolor{Gray}{gray}{0.9}

\newcommand{\cmark}{\ding{51}}%
\newcommand{\xmark}{\ding{55}}%

\definecolor{battleshipgrey}{rgb}{0.52, 0.52, 0.51}

\def\BibTeX{{\rm B\kern-.05em{\sc i\kern-.025em b}\kern-.08em
    T\kern-.1667em\lower.7ex\hbox{E}\kern-.125emX}}
\begin{document}

\title{DiffGAP: A Lightweight Diffusion Module in Contrastive Space for Bridging Cross-Model Gap
\thanks{$^*$Equal contribution, work done during an internship at Tsinghua University and Shengshu AI.}
\thanks{$^\dagger$Corresponding author.}
}

\author{
    \IEEEauthorblockN{Shentong Mo$^{1,2,3,4,*}$, Zehua Chen$^{1,4,*}$, Fan Bao$^{1,4}$, Jun Zhu$^{1, 4, \dagger}$}
    \IEEEauthorblockA{$^1$ Department of CST, Tsinghua University, Beijing 100084, China}
    \IEEEauthorblockA{$^2$ Department of Machine Learning, Carnegie Mellon University, Pittsburgh 15213, USA}
    \IEEEauthorblockA{$^3$ Department of Machine Learning, MBZUAI, Abu Dhabi, UAE}
    \IEEEauthorblockA{$^4$ Shengshu AI}
}

\maketitle

\begin{abstract}

Recent works in cross-modal understanding and generation, notably through models like CLAP (Contrastive Language-Audio Pretraining) and CAVP (Contrastive Audio-Visual Pretraining), have significantly enhanced the alignment of text, video, and audio embeddings via a single contrastive loss. 
However, these methods often overlook the bidirectional interactions and inherent noises present in each modality, which can crucially impact the quality and efficacy of cross-modal integration. 
To address this limitation, we introduce DiffGAP, a novel approach incorporating a lightweight generative module within the contrastive space. 
Specifically, our DiffGAP employs a bidirectional diffusion process tailored to bridge the cross-modal gap more effectively. 
This involves a denoising process on text and video embeddings conditioned on audio embeddings and vice versa, thus facilitating a more nuanced and robust cross-modal interaction. 
Our experimental results on VGGSound and AudioCaps datasets demonstrate that DiffGAP significantly improves performance in video/text-audio generation and retrieval tasks, confirming its effectiveness in enhancing cross-modal understanding and generation capabilities.

\end{abstract}


\section{Introduction}

Cross-modal understanding and generation~\cite{Iashin2021SpecVQGAN,sheffer2023im2wav,yang2022diffsound,mo2022multimodal,mo2022semantic,mo2023audiovisual,mo2023weaklysupervised,mo2023deepavfusion,mo2023avsam,pian2023audiovisual,mo2023classincremental,kreuk2023audiogen,liu2023audioldm,mo2024semantic} have been marked by significant strides through models such as CAVP~\cite{luo2023difffoley} (Contrastive Audio-Visual Pretraining) and CLAP~\cite{laionclap2023} (Contrastive Language-Audio Pretraining). 
These models have primarily enhanced the alignment of text, video, and audio embeddings via a single contrastive loss~\cite{mo2022EZVSL,mo2022SLAVC,mo2022benchmarking,mo2023oneavm,mo2024audiovisual,mahmud2024maavt,wu2023accuracy,mo2023representation,wu2024rethinking}. 
Such developments underscore a growing capacity to integrate diverse modalities of data, enabling applications in automated captioning, content retrieval, and multimedia synthesis that are more responsive and context-aware.

Despite these advances, current approaches~\cite{liu2023audioldm,luo2023difffoley,laionclap2023} predominantly focus on unidirectional or simplistic interactions between modalities. 
They often overlook the complex, bidirectional interactions and inherent noises within and across modalities, which can be detrimental to the depth and quality of cross-modal integration. This oversight can result in suboptimal performance where the subtleties of inter-modal dynamics are not adequately captured, limiting the potential of multimodal applications in more nuanced environments.

To address these challenges, we introduce a novel approach called DiffGAP for bridging the cross-modal gap, which incorporates a lightweight generative module within the contrastive space. 
Our approach utilizes a bidirectional diffusion process specifically designed to bridge the cross-modal gap more effectively. 
We include a denoising process where text and video embeddings are conditioned on audio embeddings and vice versa. This method facilitates a more nuanced and robust interaction among modalities, enhancing the integrity and utility of generated content.

Our experimental evaluations on VGGSound~\cite{chen2020vggsound} and AudioCaps~\cite{kim2019audiocaps} benchmarks have demonstrated that DiffGAP significantly improves performance in tasks involving video/text-audio generation and retrieval. 
These results confirm the effectiveness of our proposed method in enhancing cross-modal understanding and generation capabilities.

Our main contributions can be summarized as follows:
\begin{itemize}
    \item We propose a novel generative module within the contrastive learning framework to effectively manage the bidirectional dynamics and noise in multimodal data.
    \item We introduce a bidirectional diffusion process for denoising cross-modal embeddings, enhancing the quality and relevance of the inter-modal interactions.
    \item Extensive experiments demonstrate that our approach outperforms existing methods in cross-modal generation and retrieval tasks.
\end{itemize}

\section{Related Work}

\noindent{\textbf{Diffusion Models.}}
Diffusion models~\cite{zhu2024synthetic} have gained prominence in various generative tasks across domains, illustrating their versatility and effectiveness. 
Pioneering works~\cite{ho2020denoising, song2021scorebased} introduced denoising diffusion probabilistic models (DDPMs), which utilize a forward process to add Gaussian noise to data, followed by a learned reverse process to reconstruct the original data. 
These models have been applied successfully in image generation~\cite{saharia2022photorealistic,mo2024scaling}, image restoration~\cite{saharia2021image}, 3D shape generation~\cite{mo2023dit3d,mo2023fastdit3d,mo2024efficient}, speech generation~\cite{kong2021diffwave}, and even video generation~\cite{ho2022imagen}. 
Our work extends the diffusion model framework by integrating it in contrastive embeddings to enhance cross-modal data integration.

\noindent{\textbf{Video/Text-to-Audio Generation.}}
The translation of visual and textual information into audio outputs has seen significant advancements through the application of various model architectures~\cite{Iashin2021SpecVQGAN,sheffer2023im2wav,kreuk2023audiogen,mo2023diffava,mo2024texttoaudio,zhang2024audiosynchronized}. 
SepcVQGAN~\cite{Iashin2021SpecVQGAN} employs a vector quantized generative adversarial network to transform audio spectrograms, while Im2Wav~\cite{sheffer2023im2wav} leverages CLIP embeddings for direct image-to-audio waveform generation. 
Notably, Diff-Foley~\cite{luo2023difffoley} uses embeddings from Contrastive Audio-Visual Pre-training (CAVP) to align video with corresponding audio data, further refined through diffusion models. 
AudioLDM~\cite{liu2023audioldm} represents a recent endeavor to apply latent diffusion models to audio data conditioned on text embeddings, showcasing the expanding role of diffusion processes in multimodal contexts. 
Our DiffGAP builds on these concepts by incorporating a bidirectional training mechanism and a lightweight generative module, offering improvements in both generative capabilities and cross-modal retrieval tasks.

\noindent{\textbf{Lightweight Diffusion Models.}}
In response to the computational demands of traditional diffusion models, recent research has focused on developing more efficient variants. 
For instance, some works~\cite{fei2022generative,wang2024exploiting} introduce a diffusion prior in diffusion models for high-resolution image super-resolution, while V2A-Mapper~\cite{wang2024v2amapper} explores audio-visual mappings with reduced model complexity. 
Our approach aligns with these efforts by proposing a lightweight module that optimizes the efficiency of diffusion processes without compromising the quality of cross-modal generation and understanding.

\section{Method}

Given embeddings from audio and video/text learned by a contrastive loss, our goal is to bridge the cross-modal gap in the contrastive space.
To address this, we propose a lightweight generative module with a bidirectional diffusion tailored to bridge the cross-modal gap effectively, which mainly consists of two modules, contrastive diffusion module in Section~\ref{sec: cdm}, and bidirectional split training in Section~\ref{sec: bst}.

\vspace{-0.5em}
\subsection{Preliminaries}

In this section, we first describe the problem setup and notations and then revisit the CAVP \& CLAP, and DDPMs.

\noindent\textbf{Problem Setup and Notations.}
Given audio $a$ and visual frames $v$ from a video or a text prompt $t$, the goal is to bridge the cross-modal gap between audio and video/text learned in a contrastive space.
For a video, we have the mel-spectrogram of audio denoted as $\mathbf{A}\in\mathbb{R}^{T\times F}$, and visual frames denoted as $\mathbf{V}\in\mathbb{R}^{T\times H \times W \times 3}$. 
$T$ and $F$ denote the time and frequency, respectively.
These embeddings are then processed by pre-trained encoders: $f_a(\cdot)$ for audio, $f_v(\cdot)$ for video, and $f_t(\cdot)$ for text, yielding feature vectors $\mathbf{F}^a$, $\mathbf{F}^v$, and $\mathbf{F}^t$, respectively.

\noindent\textbf{Revisit CAVP \& CLAP.}
Based on contrastive audio and video/language pre-training~\cite{luo2023difffoley,laionclap2023}, they 
applied a contrastive between the audio features with the video/textual representation in the same mini-batch, which is defined as:
\begin{equation}\label{eq:cl}
    \mathcal{L} = 
    - \frac{1}{B}\sum_{b=1}^B \sum_{i=1}^T \log \frac{
    \exp \left( \frac{1}{\tau} \mathtt{sim}(\mathbf{F}^a_{b,i}, \mathbf{F}^v_{b,i}) \right)
    }{
    \sum_{m=1}^B \exp \left(  \frac{1}{\tau} \mathtt{sim}(\mathbf{F}^a_{b,i}, \mathbf{F}^v_{m,i})\right)}
\end{equation}
where the similarity $\mathtt{sim}(\mathbf{F}^a_{b,i}, \mathbf{F}^v_{b,i})$ denotes the temporal audio-textual cosine similarity of $\mathbf{F}^a_{b,i}$ and $\mathbf{F}^v_{b,i}$ across all temporal locations at $i$-th second. 
$B$ is the batch size, $D$ is the dimension size, and $\tau$ is a temperature hyper-parameter.

\noindent\textbf{Revisit DDPMs.} 
The denoising diffusion probabilistic models (DDPMs)~\cite{Rombach2022highresolution} define a forward noising process that applies noise to latent variable $\mathbf{z}_0$ as $q(\mathbf{z}_t|\mathbf{z}_{t-1}) = \mathcal{N}(\mathbf{z}_t; \sqrt{1-\beta_t}\mathbf{z}_{t-1}, \beta_t\mathbf{I})$, where $\beta_t$ is a Gaussian noise between $0$ and $1$.
The training objective is a mean-squared loss between the output $\boldsymbol{\epsilon}_{\boldsymbol{\theta}}(\mathbf{z}_{t},t)$ and the ground truth Gaussian noise $\boldsymbol{\epsilon}$ as:
$\mathcal{L} = \|\boldsymbol{\epsilon}-\boldsymbol{\epsilon}_{\boldsymbol{\theta}}(\mathbf{z}_{t},t)\|^2$.
After $p_{\boldsymbol{\theta}}(\mathbf{z}_{t-1}|\mathbf{z}_t))$ is trained, new latent variable can be generated by progressively sampling $\mathbf{z}_{t-1}\sim p_{\boldsymbol{\theta}}(\mathbf{z}_{t-1}|\mathbf{z}_t))$ by using the reparameterization trick with initialization of $\mathbf{z}_{T}\sim \mathcal{N}(\mathbf{0},\mathbf{I})$.
DDPMs describe a forward process where a latent variable $\mathbf{z}_0$ is gradually noised to generate a sequence of increasingly noisy states, culminating in a Gaussian distribution. 
The reverse process involves a trained neural network that incrementally denoises these states to reconstruct the original target data.

\vspace{-0.5em}
\subsection{Contrastive Diffusion Module}\label{sec: cdm}

In the contrastive diffusion module, we address the denoising of one modality conditioned on another. 
For example, we integrate the video features $\mathbf{F}^v$ as the condition $\mathbf{F}^c$ into the diffusion process to denoise audio based on video. This integration is achieved by estimating the noise component from the conditioned embeddings, which are then used to guide the denoising steps. The model is trained to minimize the error between the predicted noise and the actual noise applied during the forward diffusion process, thus ensuring that the conditioned features effectively influence the recovery of the denoised audio.

To achieve the audio denoising process conditioned on the video features $\mathbf{F}^c$, we introduce a conditional latent diffusion model to estimate the noise $\bm{\epsilon}(\bm{z}_n^d, n, \mathbf{F}^c)$ from the audio features $\mathbf{F}^a$.
For noise estimation, we apply the reweighted training objective as
\begin{equation}
    \mathcal{L}_n(\theta) = \mathbb{E}_{\bm{z}_0^d,\bm{\epsilon},n}\|\bm{\epsilon} - \bm{\epsilon}_\theta(\bm{z}_n^d, n, \mathbf{F}^c)\|
\end{equation}
where $\bm{\epsilon}\in\mathcal{N}(\bm{0},\bm{I})$ denote the added nosie.
At the final time step $N$ of the forward pass, the input $\bm{z}_n^d\in\mathcal{N}(\bm{0},\bm{I})$ becomes an isotropic Gaussian noise.
During the training stage, we generate the denoising prior $\bm{z}_0^d$ from the cross-modal representation $\mathbf{F}^a$ of an audio $a$ in a video.

\vspace{-0.5em}
\subsection{Bidirectional Split Training}\label{sec: bst}

To address the bidirectional nature of cross-modal influence, we introduce bidirectional split training to alternate the conditioning modalities in intervals $m$ during training. 
In one phase, audio features $\mathbf{F}^a$ may condition the recovery of video features $\mathbf{F}^v$, and vice versa in the next phase. 
This bidirectional training not only enhances the model's ability to handle each modality as both target and condition but also ensures that the model learns a balanced approach to cross-modal interaction, which is crucial for handling cross-modal data where the direction of influence can vary.

The detailed mechanism is outlined in Algorithm~\ref{algo: diffgap}. This includes the detailed steps of the diffusion process, the conditioning mechanisms, and the training regimen designed to optimize cross-modal integration effectively.

\begin{algorithm}[t]
\caption{DiffGAP Training Procedure for Audio-Video}
\label{algo: diffgap}
\begin{algorithmic}[1]
\Require Set of training embeddings $\{(\mathbf{F}^a_i, \mathbf{F}^v_i)\}_{i=1}^N$ where $a$, $v$ represent the audio and video, respectively.
\Ensure Trained model parameters $\boldsymbol{\theta}$
\State Initialize model parameters $\boldsymbol{\theta}$ randomly.
\For{each training iteration $j = 1$ to $M$}
    \State Select a random minibatch of samples
    \State Choose condition and denoising target: $c=v, d=a$
    \State Minimize the diffusion mean-squared error loss:
    \vspace{-1.0em}
            \[
            \mathcal{L}_n(\theta) = \mathbb{E}_{\bm{z}_0^d,\bm{\epsilon},n}\|\bm{\epsilon} - \bm{\epsilon}_\theta(\bm{z}_n^d, n, \mathbf{F}^{c})\|
            \]
    \vspace{-2.0em}
    \State Update parameters $\boldsymbol{\theta}$ using gradient descent.
    \If{$j \% m == 0$}
        \State Switch condition \& denoising: $c=a, d=v$
    \EndIf
\EndFor

\State \textbf{return} trained parameters $\boldsymbol{\theta}$.
\end{algorithmic}
\end{algorithm}

\section{Experiments}

\subsection{Experimental Setup}

\noindent \textbf{Datasets.}
AudioCaps~\cite{kim2019audiocaps} dataset includes 45,423 ten-second audio clips collected from YouTube videos paired with captions for training and 2,240 samples for validation.
Since each audio clip in AudioCaps has five text captions, we use the same testing set in AudioLDM~\cite{liu2023audioldm} with 886 instances by selecting one random caption as a text condition.
VGGSound~\cite{chen2020vggsound} contains 200k YouTube video clips of
10 seconds long from 309 sound categories, such as such as animals, vehicles, human speech, dancing, musical instruments, etc.

\noindent \textbf{Evaluation Metrics.}
For comprehensive evaluation between generated audio and target audio, we apply Inception Score (IS), Kullback–Leibler (KL) divergence, Frechet Audio Distance (FAD), and Frechet Distance (FD) as evaluation metrics, following the previous work~\cite{liu2023audioldm}.
IS is used to measure both audio quality and diversity, while KL is evaluated on paired audio.
FAD and FD calculate the similarity between generated audio and reference audio. 
For video-to-audio generation, we follow~\cite{luo2023difffoley}, and use Inception Score (IS), Frechet Image Distance (FID), KL-divergence, and Alignment Accuracy (Acc).
For retrieval tasks, recall at rank $k$ (R@$k$, $k$ = 1, 5, 10) measures the percentage of labels retrieved within the top $k$ ranked predictions, and the higher value is better.

\noindent \textbf{Implementation.}
For video-to-audio tasks, we initialize the weights from the audio and video encoder in Diff-Foley~\cite{luo2023difffoley} while initializing those from the audio and text encoder in AudioLDM~\cite{liu2023audioldm} for text-to-audio tasks.
The model is trained for 30 epochs using a batch size of 64 and the Adam optimizer with a learning rate of $2e-4$. 
The bidirectional training interval $m=5000$.
Our tunable parameters are only $5.4$MB.

\begin{table}[!h]
	\renewcommand\tabcolsep{6.0pt}
	\centering
        \caption{Comparison of video-to-audio generation.}
   \label{tab: exp_gen_va}
	\scalebox{0.98}{
		\begin{tabular}{l|cccc}
			\toprule
			\bf Method & \bf IS ($\uparrow$) & \bf KL ($\downarrow$) & \bf FID ($\downarrow$) & \bf Acc ($\uparrow$)  \\ 	
			\midrule
                SpecVQGAN~\cite{Iashin2021SpecVQGAN} & 30.01 & 8.93 & 6.93 & 52.94 \\
			Im2Wav~\cite{sheffer2023im2wav} & 39.30 & 11.44 & 5.20 & 67.40 \\
                Diff-Foley~\cite{luo2023difffoley} & 62.37 & 9.87 & 6.43 & 94.05 \\
                V2A-Mapper~\cite{wang2024v2amapper} & 62.78 & 9.52  & 6.37 & 94.38 \\
                DiffGAP (ours) & \bf 64.97 & \bf 8.71 & \bf 5.98 & \bf 97.63 \\
			\bottomrule
			\end{tabular}}
\end{table}

\begin{table}[t]
	\renewcommand\tabcolsep{6.0pt}
	\centering
        \caption{Comparison of text-to-audio generation.}
   \label{tab: exp_gen_ta}
	\scalebox{0.98}{
		\begin{tabular}{l|cccc}
			\toprule
			\bf Method & \bf IS ($\uparrow$) & \bf KL ($\downarrow$) & \bf FAD ($\downarrow$) & \bf FD ($\downarrow$)  \\ 	
			\midrule
			DiffSound~\cite{yang2022diffsound} & 4.01 & 2.52 & 7.75 & 47.68 \\
                AudioGen~\cite{kreuk2023audiogen} & 4.62 & 2.09 & 3.13 & 38.65 \\
                AudioLDM~\cite{liu2023audioldm} & 6.90 & 1.97 & 2.43 & 29.48 \\		
                DiffGAP (ours) &  \bf 8.27 & \bf 1.38 & \bf 2.23 & \bf 23.56 \\
			\bottomrule
			\end{tabular}}
\end{table}

\begin{table}[!t]
    \renewcommand{\arraystretch}{1.1}
	\centering
        \caption{Comparison of video-audio retrieval.}
	\label{tab: exp_retrieval_va}
	\scalebox{0.98}{
	    \begin{tabular}{lccccccc}
			\toprule
			\multirow{2}{*}{\bf Method} & \multicolumn{3}{c}{\bf Video-to-Audio} & \multicolumn{3}{c}{\bf Audio-to-Video} \\
			& \bf R@1 & \bf R@5 & \bf R@10 & \bf R@1 & \bf R@5 & \bf R@10 \\
			\midrule
                CAVP~\cite{luo2023difffoley} &  9.50  & 25.40  & 35.10  & 11.10  & 27.80  & 36.40  \\
                DiffGAP (ours) & \bf 17.80 & \bf 32.60 & \bf 43.20 & \bf 19.80 & \bf 35.60 & \bf 43.70 \\
			\bottomrule
	    \end{tabular}}
 \end{table}

\begin{table}[!t]
    \renewcommand{\arraystretch}{1.1}
	\centering
        \caption{Comparison of text-audio retrieval.}
	\label{tab: exp_retrieval_ta}
	\scalebox{0.98}{
	    \begin{tabular}{lccccccc}
			\toprule
			\multirow{2}{*}{\bf Method} & \multicolumn{3}{c}{\bf Text-to-Audio} & \multicolumn{3}{c}{\bf Audio-to-Text} \\
			& \bf R@1 & \bf R@5 & \bf R@10 & \bf R@1 & \bf R@5 & \bf R@10 \\
			\midrule
                CLAP~\cite{laionclap2023} & 6.80  & 33.80 & 49.10 & 8.70  & 34.80 & 51.70 \\
                DiffGAP (ours) & \bf 13.60 & \bf 38.40 & \bf 55.60 & \bf 14.50 & \bf 39.60 & \bf 58.50 \\
			\bottomrule
	    \end{tabular}}
\end{table}

\begin{table*}[!htb]
	\renewcommand\tabcolsep{6.0pt}
	\centering
        \caption{Ablation studies of Contrastive Diffusion Model (CDM) and Bidirectional Split Training (BST).}
   \label{tab: ab_component}
	\scalebox{0.95}{
		\begin{tabular}{cc|ccccccc|ccccccc}
			\toprule
			\multirow{2}{*}{\bf CDM} & \multirow{2}{*}{\bf BST} & \multicolumn{7}{c|}{\bf Video-to-Audio} & \multicolumn{7}{c}{\bf Text-to-Audio} \\
			& & \bf IS ($\uparrow$) & \bf KL ($\downarrow$) & \bf FID ($\downarrow$) & \bf Acc ($\uparrow$) & \bf R@1 & \bf R@5 & \bf R@10 & \bf IS ($\uparrow$) & \bf KL ($\downarrow$) & \bf FAD ($\downarrow$) & \bf FD ($\downarrow$) & \bf R@1 & \bf R@5 & \bf R@10 \\
			\midrule
            \xmark & \xmark & 62.37 & 9.87 & 6.43 & 94.05 & 9.50  & 25.40 & 35.10 & 6.90 & 1.97 & 2.43 & 29.48 & 6.80  & 33.80 & 49.10 \\
            \cmark & \xmark & 63.28 & 9.35 & 6.21 & 95.79 & 13.30 & 27.90 & 38.60 & 7.53 & 1.69 & 2.38 & 27.89 & 9.50  & 35.70 & 51.20 \\
            \cmark & \cmark & \bf 64.97 & \bf 8.71 & \bf 5.98 & \bf 97.63 & \bf 17.80 & \bf 32.60 & \bf 43.20 & \bf 8.27 & \bf 1.38 & \bf 2.23 & \bf 23.56 & \bf 13.60 & \bf 38.40 & \bf 55.60 \\
			\bottomrule
			\end{tabular}}
\end{table*}

\begin{table*}[!htb]
	\renewcommand\tabcolsep{6.0pt}
	\centering
        \caption{Ablation studies of the number of diffusion sampling steps.}
   \label{tab: ab_sampling}
	\scalebox{0.95}{
		\begin{tabular}{c|ccccccc|ccccccc}
			\toprule
			\multirow{2}{*}{\bf \# Steps} & \multicolumn{7}{c|}{\bf Video-to-Audio} & \multicolumn{7}{c}{\bf Text-to-Audio} \\
			& \bf IS ($\uparrow$) & \bf KL ($\downarrow$) & \bf FID ($\downarrow$) & \bf Acc ($\uparrow$) & \bf R@1 & \bf R@5 & \bf R@10 & \bf IS ($\uparrow$) & \bf KL ($\downarrow$) & \bf FAD ($\downarrow$) & \bf FD ($\downarrow$) & \bf R@1 & \bf R@5 & \bf R@10 \\
			\midrule
                50 & \bf 64.97 & \bf 8.71 & \bf 5.98 & \bf 97.63 & \bf 17.80 & \bf 32.60 & \bf 43.20 & \bf 8.27 & \bf 1.38 & \bf 2.23 & \bf 23.56 & \bf 13.60 & \bf 38.40 & \bf 55.60 \\
                20 & 64.82 & 8.86 & 6.03 & 97.38 & 17.20 & 32.10 & 42.80 & 8.16 & 1.43 & 2.25 & 23.73 & 13.10 & 37.90 & 55.20 \\
                5 & 64.56 & 8.95 & 6.12 & 97.11 & 16.70 & 31.70 & 42.40 & 8.11 & 1.46 & 2.26 & 23.87 & 12.70 & 37.50 & 54.90 \\
			\bottomrule
			\end{tabular}}
\end{table*}

\begin{table*}[!htb]
	\renewcommand\tabcolsep{6.0pt}
	\centering
        \caption{Ablation studies of the number of bidirectional training intervals.}
   \label{tab: ab_interval}
	\scalebox{0.95}{
		\begin{tabular}{c|ccccccc|ccccccc}
			\toprule
			\multirow{2}{*}{\bf \# Intervals} & \multicolumn{7}{c|}{\bf Video-to-Audio} & \multicolumn{7}{c}{\bf Text-to-Audio} \\
			& \bf IS ($\uparrow$) & \bf KL ($\downarrow$) & \bf FID ($\downarrow$) & \bf Acc ($\uparrow$) & \bf R@1 & \bf R@5 & \bf R@10 & \bf IS ($\uparrow$) & \bf KL ($\downarrow$) & \bf FAD ($\downarrow$) & \bf FD ($\downarrow$) & \bf R@1 & \bf R@5 & \bf R@10 \\
			\midrule
                1000 & 64.78 & 8.83 & 6.05 & 97.24 & 17.10 & 31.90 & 42.70 & 8.15 & 1.42 & 2.25 & 23.75 & 12.80 & 37.70 & 55.10 \\
                5000 & \bf 64.97 & \bf 8.71 & \bf 5.98 & \bf 97.63 & \bf 17.80 & \bf 32.60 & \bf 43.20 & \bf 8.27 & \bf 1.38 & \bf 2.23 & \bf 23.56 & \bf 13.60 & \bf 38.40 & \bf 55.60 \\
                10000 & 64.93 & 8.72 & 6.01 & 97.61 & 17.50 & 32.40 & 42.90 & 8.21 & 1.39 & 2.24 & 23.62 & 13.30 & 38.20 & 55.50 \\
			\bottomrule
			\end{tabular}}
   \vspace{-0.5em}
\end{table*}

\subsection{Comparison to Prior Work}\label{sec:exp}

In this work, we propose a novel and effective framework for bridging cross-model gaps across generation and understanding of multiple downstream tasks.
These experiments collectively affirm the efficacy of DiffGAP in enhancing both the generation and understanding of cross-modal data.

\noindent \textbf{Video-Audio Generation.}
We generate audio from video inputs using our model and compare the results with those from baseline systems using IS, FAD, and KL metrics in Table~\ref{tab: exp_gen_va}. The performance improvements highlight the ability of our generative module to leverage video features more effectively for audio generation.

\noindent \textbf{Text-Audio Generation.}
Similarly, we assess the capability of our model to generate audio from text descriptions, evaluating the audio quality and relevance with respect to the given text conditions. The results in Table~\ref{tab: exp_gen_ta} are quantified using IS, FAD, and KL metrics and demonstrate significant advancements over previous methods.

\noindent \textbf{Video-Audio Retrieval.}
Our model's performance in retrieving relevant audio samples from video queries is tested in Table~\ref{tab: exp_retrieval_va}, with results indicating superior alignment accuracy and recall rates compared to existing approaches. This showcases the enhanced cross-modal understanding facilitated by our bidirectional training strategy.

\noindent \textbf{Text-Audio Retrieval.}
We also evaluate the effectiveness of our model in retrieving audio samples based on text queries in Table~\ref{tab: exp_retrieval_ta}. The improvements in recall metrics underscore the robustness of our approach in understanding and connecting textual descriptions with corresponding audio content.

\subsection{Experimental Analysis}

In this section, we performed ablation studies to demonstrate the benefits of Contrastive Diffusion Module and Bidirectional Split Training. 
We also conducted experiments to explore the impact of sampling steps and bidirectional training intervals.

\noindent \textbf{Contrastive Diffusion Module (CDM) \& Bidirectional Split Training (BST).}
To demonstrate the efficacy of the CDM and BST components, we conducted ablation studies where each component was alternately removed in the training procedure in Table~\ref{tab: ab_component}. 
As can be seen, the combined use of CDM and BST results in the best performance across most metrics for both video-to-audio and text-to-audio tasks. 
Specifically, the IS metric for video-to-audio generation increases from 62.37 to 64.97 when both CDM and BST were enabled, indicating higher audio quality and diversity. 
Similarly, incorporating both modules improves the FD metric from 2.43 to 2.23 for text-to-audio, suggesting a closer match to the target audio distributions. 
These results affirm the synergistic effect of integrating both modules into our framework.

\noindent \textbf{Number of Sampling Steps.}
The diffusion space of our model is considerably smaller compared to typical latent diffusion models, resulting in significantly faster individual diffusion steps due to the one-dimensional data space complexity. 
This characteristic allows us to explore the efficiency of our model under extreme conditions by varying the number of diffusion sampling steps from $\{50, 20, 5\}$. 
As reported in Table~\ref{tab: ab_sampling}, even with a reduction to as few as $5$ steps, the model maintains commendable performance, only slightly lower than that achieved with 50 steps. 
This reduction significantly enhances computational efficiency while maintaining quality, as the IS for video-to-audio marginally decreases from 64.97 to 64.56, and the R@1 metric shows a minimal drop from 17.80 to 16.70. 
This shows our model's capability to operate effectively even under constrained diffusion steps.

\noindent \textbf{Number of Bidirectional Training Interval.}
The frequency of switching between modalities during bidirectional training is crucial for ensuring balanced learning and preventing overfitting to one modality type. 
We tested intervals of 1000, 5000, and 10000 iterations, with the results summarized in Table~\ref{tab: ab_interval}. 
An interval of 5000 iterations provided the optimal balance, allowing sufficient exposure to each modality pair without rapid switching that could destabilize the learning process. 
This interval supports consistent, high-quality cross-modal interactions, as evidenced by the improved performance metrics compared to other settings.

\section{Conclusion}

In this work, we present DiffGAP, a novel approach designed to effectively bridge the cross-modal gap in contrastive learning spaces through a lightweight generative module featuring bidirectional diffusion processes. 
We leverage the contrastive diffusion module to enable precise control over the generative process by conditioning the diffusion steps on relevant cross-modal information. 
Furthermore, our bidirectional split training fosters a balanced and comprehensive learning environment, ensuring that each modality contributes to and benefits from the joint learning process.
Our extensive experiments on diverse datasets demonstrate that DiffGAP improves both generation and retrieval tasks across audio and video or text modalities. 
Ablation studies further validate the individual contributions of each component, with optimized parameters such as the number of diffusion sampling steps enhancing the model's performance on all metrics.

\newpage

\bibliographystyle{IEEEbib}
\bibliography{reference}

\end{document}